\documentclass[12pt]{article}
\usepackage{spconf,amsmath,epsfig}
\usepackage{tikz}
\usepackage{multirow}
\usepackage{makecell}
\usepackage{pgfplots}
\pgfplotsset{compat=1.17}
\usetikzlibrary{shapes.geometric, arrows}
\usepackage{tcolorbox}
\tcbuselibrary{breakable}

\usepackage[a4paper,margin=2cm]{geometry}

\tikzstyle{process} = [rectangle, rounded corners, minimum width=3cm, minimum height=1cm, text centered, draw=black, fill=blue!20]
\tikzstyle{arrow} = [thick,->,>=stealth]
\usepackage{lipsum}
\usepackage{booktabs}
\usepackage{subcaption}
\usepackage{rotating}
\usepackage{tcolorbox}
\usepackage{enumitem}
\usepackage[colorinlistoftodos,prependcaption,textsize=normalsize]{todonotes}
\usepackage{float}
\usepackage[commandnameprefix=always]{changes}
\tcbuselibrary{listingsutf8}
\providecommand{\doi}[1]{doi: {\footnotesize \href{http://dx.doi.org/#1}{\path{#1}}}}

\usepackage[pdftex=true,breaklinks=true,hidelinks=true,colorlinks=true,citecolor=blue]{hyperref}

\usepackage[square,numbers]{natbib}

\title{Behavioural self-monitoring for autonomous earth observation agents: The attribute reliability monitor (ARM)}
\name{
\begin{tabular}{c}
Alejandro D. Mousist
\end{tabular}
}

\address{Thales Alenia Space España\\
C/ Einstein 7 (PTM)\\
28760 Tres Cantos, Madrid, Spain\\
alejandro.mousist@thalesaleniaspace.com}

\begin{document}
\raggedbottom
%
\maketitle
%
\begin{abstract}
Language and vision language models can support autonomous systems, but changing operating conditions may compromise the reliability of their semantic outputs. This work introduces the Attribute Reliability Monitor (ARM), a framework that accumulates evidence of suspicion over time to assess individual semantic attributes. ARM accommodates different detectors and temporal monitoring methods, while leaving intervention policies to the consuming system. An initial implementation was evaluated using a specialized vision language model developed for satellite thermal infrared imagery. Monitoring followed a black box approach, using only observable responses without access to internal model states, confidence scores, or external reference answers. Under controlled input degradation, disagreement between responses provided evidence for selective suspension of attributes. The results illustrate how external behavioral monitoring can inform continued use of model outputs, while distinguishing operational suspicion from semantic correctness.
\end{abstract}

%

\section{INTRODUCTION}
\label{sec:introduction}
Large language models and vision language models are enabling new approaches to systems that interpret observations, use tools, and revise plans in response to intermediate results. In agent architectures, these models can support perception, planning, and coordination through natural language interfaces and access to external tools \cite{embodied_agentic_ai}. This creates opportunities for autonomous systems that must respond to changing conditions and situations not explicitly anticipated during their design.

Deploying such models near the source of observations could allow remote systems to interpret local conditions and adapt their actions without requesting human intervention at every step. For space missions, this prospect is particularly relevant when communication delays or limited contact opportunities make continuous supervision difficult. Initial demonstrations include VLM inference on imagery acquired by an Earth observation satellite \cite{navi_orbital}, LLM supervision of thermal control aboard the International Space Station \cite{astrea}, and autonomous sampling and fault recovery on a terrestrial testbed for planetary exploration \cite{ai_space_cortex}. These developments suggest opportunities for broader mission autonomy, subject to the computational resources and operational constraints of each platform.

The use of language models in autonomous decision making also introduces a monitoring challenge. A model may generate unsupported or incorrect information while presenting it fluently and convincingly \cite{xu_hallucination}. If an agent uses such outputs to select tools, revise a plan, or interpret an observation, errors may propagate into subsequent actions. Moreover, a model validated under one set of conditions may encounter different environments or degraded sensor inputs during operation. These changes may affect individual semantic outputs differently, making a single assessment of the model insufficient for deciding which outputs remain suitable for use.

Autonomous systems would therefore benefit from mechanisms that monitor the behavior of their models over time and provide evidence for selective intervention. Such mechanisms should distinguish between the semantic attributes on which downstream decisions depend, rather than requiring the entire model to be accepted or rejected as a whole. When an attribute becomes suspect, the consuming system could seek an alternative model or tool, request further verification, or restrict the operations that depend on it.

This paper introduces the Attribute Reliability Monitor (ARM), a framework that accumulates evidence of suspicion for individual semantic attributes and maintains an operational assessment of their reliability. ARM does not prescribe a particular detector or intervention policy, allowing different evidence sources and temporal monitoring methods to be used. An initial implementation is evaluated with a thermal infrared vision language model under controlled input degradation. The implementation measures disagreement between responses to alternative prompt formulations and monitors its evolution using cumulative sum control charts. The experiment examines how this evidence can support selective suspension of model outputs without supplying reference labels to the monitor; it does not establish semantic correctness or validate autonomous mission operation.

\section{Related Work}
\label{sec:sota}
Hallucinations occur when a model generates information that is factually incorrect, unsupported, or inconsistent with the supplied context. Xu et al. \cite{xu_hallucination} show, under a formal formulation, that hallucinations cannot be completely eliminated when LLMs are used as general problem solvers. This motivates mechanisms for detecting unreliable outputs during operation.

Black box hallucination detection methods do not require access to token probabilities or internal model states. Some approaches rely on external knowledge bases or auxiliary models, while methods described as zero resource avoid external factual references and use observable model outputs as evidence. This reduces dependence on external databases and continuous connectivity, which is relevant to onboard systems, although repeated inference can still incur substantial computational cost.

SelfCheckGPT \cite{selfcheckgpt} samples multiple responses to the same query and estimates factuality from their informational consistency. Its underlying assumption is that factual content is more likely to recur across samples, whereas hallucinated content is more likely to diverge. This assumption can fail when a model repeatedly produces the same incorrect answer.

Kong et al. \cite{bidirectional_detector} seek to mitigate this limitation by combining consistency checks from two perspectives. They compare responses sampled from the original query and reconstruct multiple queries from the response being evaluated. The reconstructed queries are then compared with the original query. This backward check may reveal discrepancies that are not apparent from agreement between responses alone. The response and query consistency scores are combined into a single hallucination detection score.

Yang et al. \cite{llm-detection} propose a trained LLM evaluator that assesses query and answer pairs using adaptive criteria and produces an interpretable score without accessing the evaluated model's internals or external references at inference. Using an additional LLM introduces computational and memory demands that may be difficult to accommodate onboard satellites with limited resources.

These approaches assess individual responses or interactions. When autonomous agents use the same model over prolonged periods, input conditions may evolve and depart from those under which the model was originally validated, a situation addressed by research on concept drift \cite{sensor_drift}. Such changes may affect the reliability of individual semantic outputs over time.

This issue is particularly relevant in multimodal systems. A VLM coupled with an imaging sensor may encounter new geographical regions, seasonal variations, or previously unseen environmental conditions. Sensor ageing and calibration changes may also alter the input distribution. These changes need not affect every semantic capability equally, motivating a temporal assessment for each attribute. The response detection methods discussed above do not explicitly maintain such an assessment across successive observations.

At a higher system level, changes in model behavior can affect the agents that depend on its outputs. Rath \cite{agent_drift} examines behavioral degradation in agents over extended interactions, providing a complementary motivation for monitoring beyond isolated responses. This concerns evolution during agent interaction, whereas the experiment in this study examines changes induced by degradation of the visual input.

Together, these considerations motivate mechanisms that accumulate behavioral evidence over time and maintain an operational assessment of individual semantic capabilities.

\section{Attribute Reliability Monitor}
\label{sec:arm}

In task-oriented perception pipelines and agentic systems, outputs from language-based models can be mapped to a predefined set of structured semantic attributes, enabling downstream reasoning, monitoring, and decision-making. The reliability of these attributes may vary during operation.

The Attribute Reliability Monitor (ARM) is a detector-agnostic framework for monitoring attribute-level evidence over time and using it to assess whether each attribute output should continue to be trusted. For each semantic attribute $a$, a selected detection method produces instantaneous evidence of suspicion $e_a(t)$. ARM does not prescribe how this evidence is generated.

The evidence may take different forms depending on the selected detection method:
\begin{equation}
    e_a(t) \in \mathcal{E}_a,
\end{equation}
where $\mathcal{E}_a$ is the corresponding evidence space. For example, the evidence may be binary, continuous, or categorical.

Over time, ARM maintains a temporal memory of the evidence associated with attribute $a$:
\begin{equation}
    M_a(t) = \mathcal{U}_a\bigl(M_a(t-1), e_a(t)\bigr),
\label{eq:time}
\end{equation}
where $\mathcal{U}_a$ is a temporal update function. The memory may retain the evidence history or a summary of it, depending on the monitoring requirements.

The reliability assessment for attribute $a$ at time $t$ is obtained by applying an aggregation function $\mathcal{A}$ to this temporal memory:
\begin{equation}
    R_a(t) = \mathcal{A}\bigl(M_a(t)\bigr).
\label{eq:reliability}
\end{equation}
ARM does not prescribe a specific aggregation function or output representation. Depending on the application, $R_a(t)$ may be a continuous score or a discrete operational state, such as trusted or suspended. Unless calibrated against ground-truth outcomes, it represents ARM's operational assessment of reliability rather than a probability that the attribute is correct.

\begin{figure}[H]
  \centering
  \includegraphics[width=0.8\columnwidth]{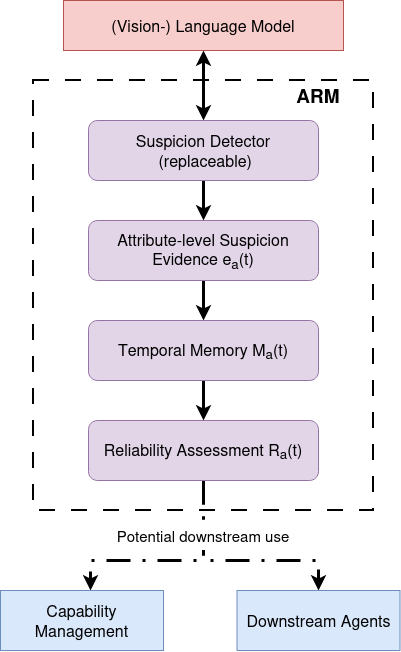}
  \caption{Conceptual architecture of the Attribute Reliability Monitor (ARM)}
  \label{fig:arm}
\end{figure}

Fig.~\ref{fig:arm} illustrates the conceptual ARM workflow.

The reliability assessment produced by ARM concerns a particular model's semantic output in its operating context. Based on $R_a(t)$ and its own requirements, a consuming system may continue to use that output, select an alternative model or tool, or request further assessment. Suspending the use of an attribute from one model does not necessarily remove the corresponding system capability. ARM provides the reliability assessment; the consuming system defines the policy for acting on it.

\section{Experimental Evaluation}
\label{sec:experiment}

\subsection{Monitored Model and Dataset}
ARM was conceived to monitor language and vision language models through their external behavior. To evaluate the approach, a specialized thermal infrared model (TIR-VLM) was developed as a controlled experimental testbed. Its performance varies across semantic attributes, allowing ARM to be evaluated under different levels of input degradation.

Landsat 8 imagery was used to construct the training and evaluation data. Band 10 was selected as the sole input modality due to known calibration issues affecting Band 11.

Before visual encoding, the thermal image passes through a fixed preprocessing layer that derives six additional representations, yielding seven channels:

\begin{itemize}[noitemsep, topsep=0pt]
    \item Channel 0: Normalized thermal intensity
    \item Channel 1: Local Z-score
    \item Channel 2: Sobel gradient magnitude
    \item Channel 3: Laplacian
    \item Channel 4: Local variance
    \item Channel 5: Difference of Gaussians
    \item Channel 6: High frequency residual
\end{itemize}

\begin{figure*}[ht]
  \centering
  \includegraphics[width=\textwidth]{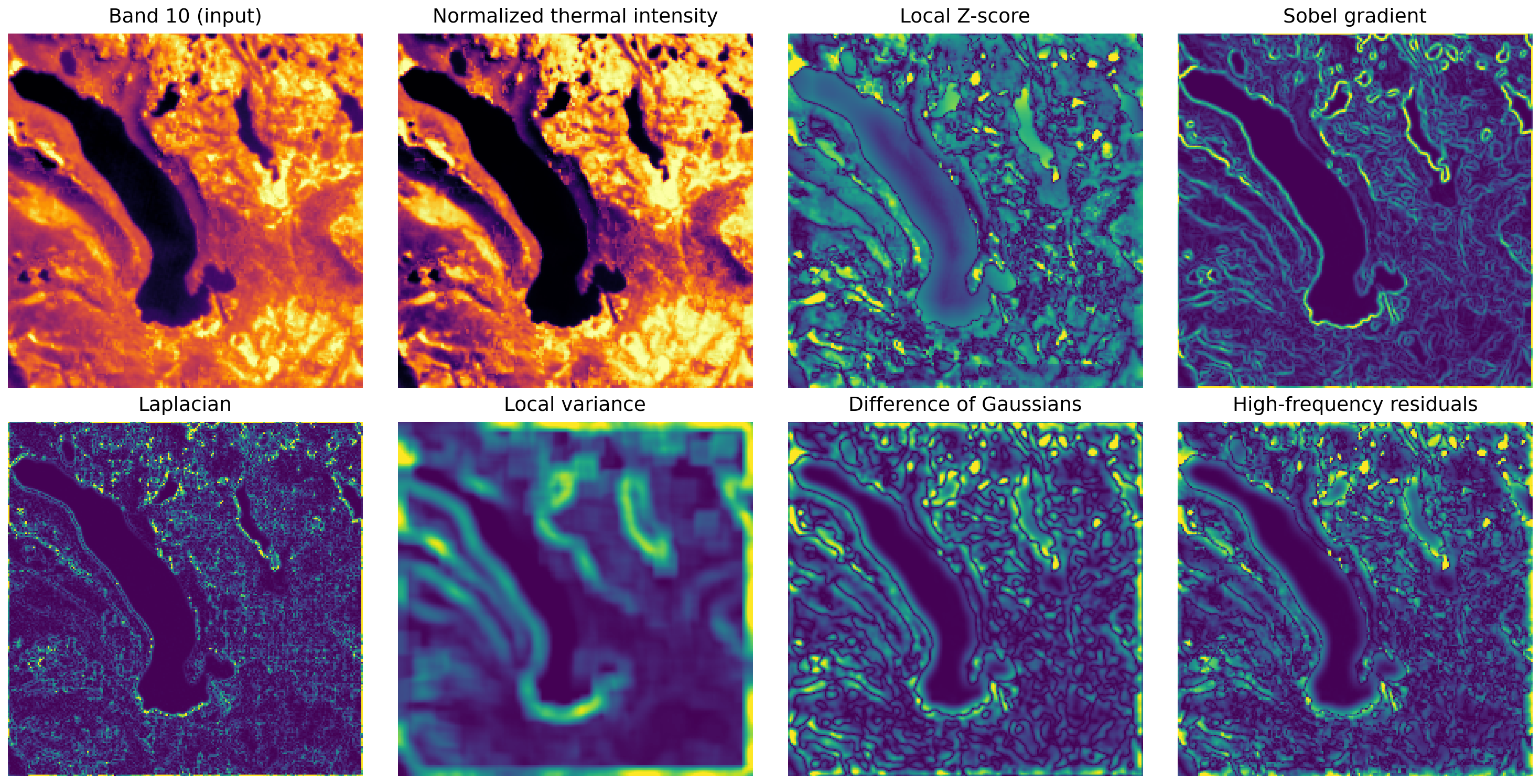}
  \caption{Representations derived from Landsat 8 thermal infrared Band 10.}
  \label{fig:augmentation}
\end{figure*}

Normalization uses fixed statistics computed from the training images. Fig. \ref{fig:augmentation} illustrates the representations derived from the original Band 10 observation.

The visual encoder is based on ConvNeXt Tiny \cite{convnext}, pretrained using a masked reconstruction objective on thermal imagery. Qwen2.5 1.5B \cite{qwen2.5} provides the language component.

For a $224 \times 224$ input, the encoder produces a final $7 \times 7$ feature map with 768 channels. Learned spatial position embeddings are added before the map is flattened into 49 visual tokens. An MLP with two linear layers and a GELU activation maps each token from 768 to the 1536 dimensions of the Qwen2.5 embedding space. The projected tokens are normalized and multiplied by a learned scale, then enclosed by two learned delimiter tokens before being passed to the language model.

During alignment, the visual encoder remains frozen. The MLP, position embeddings, scale and delimiter tokens are trained from scratch, while Qwen2.5 is adapted using Low Rank Adaptation (LoRA) \cite{lora}. No synthetic banding was applied during encoder pretraining or model alignment. The additive banding described in Section~4.3 was introduced only during evaluation.

The source data are the SPARCS \cite{SPARCS} Landsat 8 scenes, a collection of 80 subscenes, each measuring $1000 \times 1000$ pixels, with manually interpreted reference masks. The masks contain seven classes:
\begin{itemize}[noitemsep, topsep=0pt]
    \item Cloud shadow
    \item Cloud shadow over water
    \item Water
    \item Ice/snow
    \item Land
    \item Clouds
    \item Flooded
\end{itemize}
The \emph{flooded} class was excluded because it appeared in too few scenes.

For this study, 78 subscenes were used: 63 for training and 15 for validation. The split was performed at the source scene level before patch extraction. Each image was divided into nine patches of $512 \times 512$ pixels, with patch origins separated by 244 pixels and an overlap of 268 pixels between adjacent patches. This yielded 567 training patches and 135 validation patches. Each patch was then resized to $224 \times 224$ pixels for input to the visual encoder.

Each example pairs a Landsat 8 Band 10 (B10) patch with a prompt and a structured answer containing the reference coverage fractions for the retained classes and, where applicable, the dominant class.


\subsection{Experimental Instantiation of ARM}
\label{sec:experimental_instantiation}

For the experimental instantiation of ARM, TIR-VLM was queried using six fixed prompts requesting overlapping descriptions of each image:

\begin{itemize}[noitemsep, topsep=0pt]
    \item ``Present a structured account of the image composition.''
    \item ``Summarize the scene depicted in this image as a whole.''
    \item ``Provide a structured description of the full scene shown here.''
    \item ``Describe class coverage throughout the image along with the leading class.''
    \item ``Describe how the scene is composed, including its predominant class.''
    \item ``Present the scene through its predominant class and the relative extent of the classes.''
\end{itemize}

Each prompt required the same JSON structure, containing the coverage fractions of the six attributes and the dominant class. For each attribute, inconsistency evidence was calculated as the root mean square of the differences between the $\binom{6}{2}=15$ distinct pairs of responses:
\[
e_a(t)=
\sqrt{
    \frac{1}{15}
    \sum_{1\leq i<j\leq6}
    \left[p_{a,i}(t)-p_{a,j}(t)\right]^2
},
\]
where $p_{a,i}(t)$ is the coverage fraction reported for attribute $a$ by prompt $i$ for image $t$. This measure captures disagreement between formulations and gives greater weight to large differences. Only the coverage fractions were monitored; the dominant class was not used to compute the evidence.

In this implementation, temporal monitoring uses a cumulative sum (CUSUM) control chart \cite{cusum}, a sequential technique for detecting changes in a monitored signal. A separate CUSUM monitor is maintained for each attribute to detect increases in the inconsistency evidence obtained from the six responses. 

The first pass over the 135 clean patches provided 134 valid observations, which established the mean and sample standard deviation of the evidence for each attribute. Subsequent evidence was standardized by subtracting the reference mean and dividing by its standard deviation. Each monitor starts with a zero accumulator and updates it by adding the standardized evidence minus an allowance of $k=0.25$, with a lower bound of zero. Standardized evidence above this allowance increases the accumulator, while smaller values reduce it towards zero. The accumulator therefore implements the temporal memory described in Section~\ref{sec:arm}, without requiring a fixed monitoring window.

Inconsistency evidence was computed only when all six responses contained valid coverage fractions. In each degradation experiment, two image evaluations contained an invalid response and were excluded from evidence calculation. At an invalid monitored observation, the CUSUM accumulators retained their previous values.

The decision threshold was calibrated using 2000 simulations based exclusively on responses to the 567 clean training patches from 63 source scenes. Of these patches, 565 provided valid inconsistency evidence. Each simulation sampled 15 source scenes with replacement, shuffled their patches, and repeated the resulting sequence of 135 patches for six passes. The first pass established the reference, and the remaining five were monitored. The 95th percentile of the maximum CUSUM value across all attributes and monitored steps gave a common threshold of $h\approx14.54$, targeting a joint alarm probability of 5\% under these simulated clean conditions. This threshold was fixed before evaluating either degradation scenario; each experiment established its operational reference from its own clean validation pass.

For each experiment, the adopted policy enabled all attributes after calibration and disabled an attribute at its first CUSUM threshold crossing. It remained disabled for the rest of the sequence. This policy represents one possible response to ARM's assessment; ARM itself does not prescribe deactivation. Reference labels were used separately to evaluate coverage error and were not supplied to the monitor.

\begin{figure}[ht]
  \centering
  \includegraphics[width=\columnwidth]{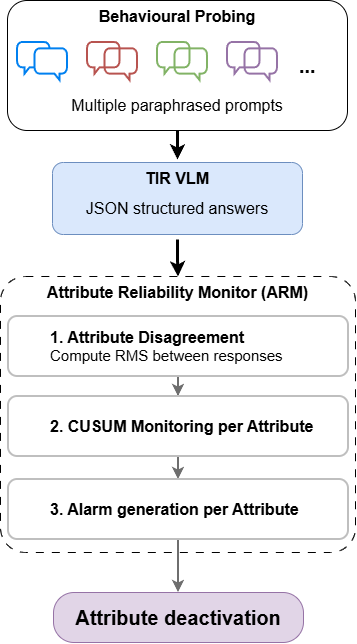}
  \caption{Behavioral probing and attribute deactivation for ARM evaluation.}
  \label{fig:experiment}
\end{figure}

Fig. \ref{fig:experiment} illustrates this implementation of ARM.

\subsection{Controlled Input Degradation}

Two controlled perturbations were selected to evaluate ARM: additive banding and Gaussian blur. These represent changes in image intensity and loss of spatial detail, respectively, inspired by effects that may affect optical instruments. They provide controlled degradation scenarios rather than physical models of specific TIRS failures.

\begin{figure*}[ht]
  \centering

  \begin{subfigure}{\textwidth}
    \centering
    \includegraphics[width=\linewidth]{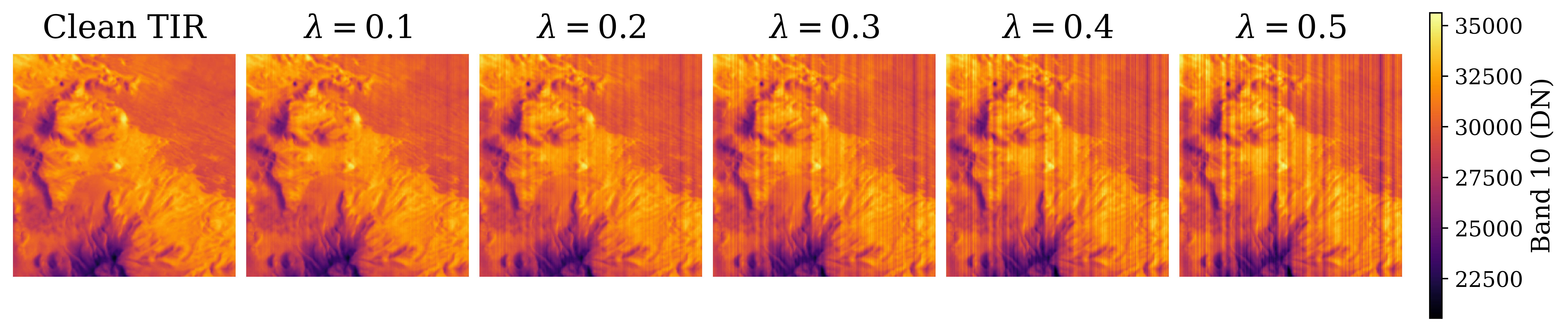}
    \caption{Additive banding.}
    \label{fig:noise_banding}
  \end{subfigure}

  \medskip

  \begin{subfigure}{\textwidth}
    \centering
    \includegraphics[width=\linewidth]{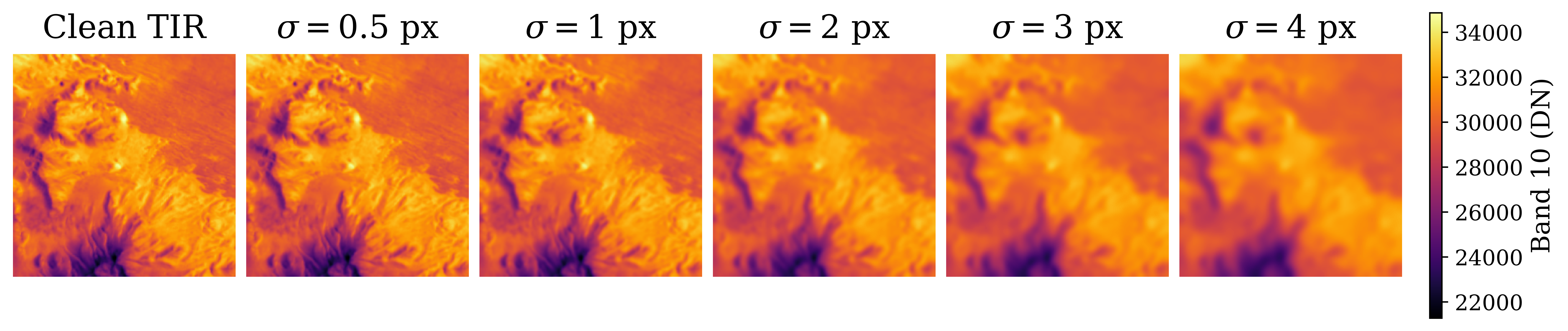}
    \caption{Gaussian blur.}
    \label{fig:noise_blur}
  \end{subfigure}

  \caption{The same TIR patch under increasing input degradation:
  (a) additive banding controlled by $\lambda$;
  (b) Gaussian blur controlled by $\sigma$, measured in pixels
  at the model input resolution.}
  \label{fig:noise}
\end{figure*}

For each experiment, the same 135 validation patches were processed in the same order at six degradation levels, yielding 810 image evaluations. Perturbations were applied to the resized $224 \times 224$ Band 10 input before preprocessing. The clean pass established the reference, and the five subsequent passes were monitored without resetting the CUSUM accumulators. The two experiments used separate accumulators and the same decision threshold calibrated from clean training observations.
Figure~\ref{fig:noise} illustrates the same patch under the six levels of each perturbation.

Responses were generated deterministically with \texttt{do\_sample=False}, so disagreement reflects differences between prompt formulations rather than random sampling. Coverage error was evaluated using mean absolute error (MAE): for each attribute and degradation level, absolute errors were averaged over the valid responses for each patch and then across patches with valid responses. MAE is reported in percentage points.

\subsubsection{Additive Banding}

Progressive banding was introduced by adding a fixed pattern of column offsets:
\begin{equation}
    \widetilde{I}_{\lambda}(r,c)
    = I(r,c) + \lambda\,\sigma_{\mathrm{ref}}\,b(c),
    \label{eq:synthetic_banding}
\end{equation}
where $I(r,c)$ is the original Band 10 value and $b(c)$ is generated from Gaussian column offsets, smoothed over five columns and normalized to zero mean and unit standard deviation. Each offset is repeated across all rows, producing vertical bands. The reference scale $\sigma_{\mathrm{ref}}$ is the median standard deviation of the clean resized Band 10 patches.

The same pattern was applied to every image at strengths $\lambda \in \{0, 0.1, 0.2, 0.3, 0.4, 0.5\}$, where $\lambda=0$ corresponds to the clean input. This perturbation is inspired by detector nonuniformity in pushbroom sensors. Synthetic banding was introduced only during evaluation and was not used during model training.

\subsubsection{Gaussian Blur}

Progressive loss of spatial detail was simulated by convolving each image with a Gaussian kernel. Its standard deviation was set to $\sigma \in \{0, 0.5, 1, 2, 3, 4\}$ pixels, measured at the $224 \times 224$ input resolution. The case $\sigma=0$ retained the clean image.

Increasing $\sigma$ smooths local intensity variations and reduces the visibility of boundaries and fine structures. The same blur strength was applied to every patch within each pass, while the reference coverage fractions remained unchanged.


\subsection{Results}
\label{sec:arm_results}

Together, the experiments illustrate selective responses to accumulated inconsistency evidence using the same monitoring method and decision threshold. The responses differ across attributes and perturbations. Increased disagreement may occur without higher overall MAE, while coverage error may increase without triggering an alarm. ARM therefore provides evidence for operational reassessment of individual outputs; its alarms and enabled states do not establish semantic correctness.

\subsubsection{Additive Banding}
Table~\ref{tab:arm_results} summarizes the results obtained as the validation patches were processed at increasing banding levels. All attributes were enabled after calibration. Ice/Snow was disabled at $\lambda=0.2$, followed by Land at $\lambda=0.4$ and Cloud shadow at $\lambda=0.5$. Water, Clouds, and Cloud shadow over water remained enabled. These noise levels identify when alarms occurred in this sequence, rather than fixed noise tolerance limits for each attribute.

\begin{figure*}[ht]
  \centering
  \includegraphics[width=\textwidth]{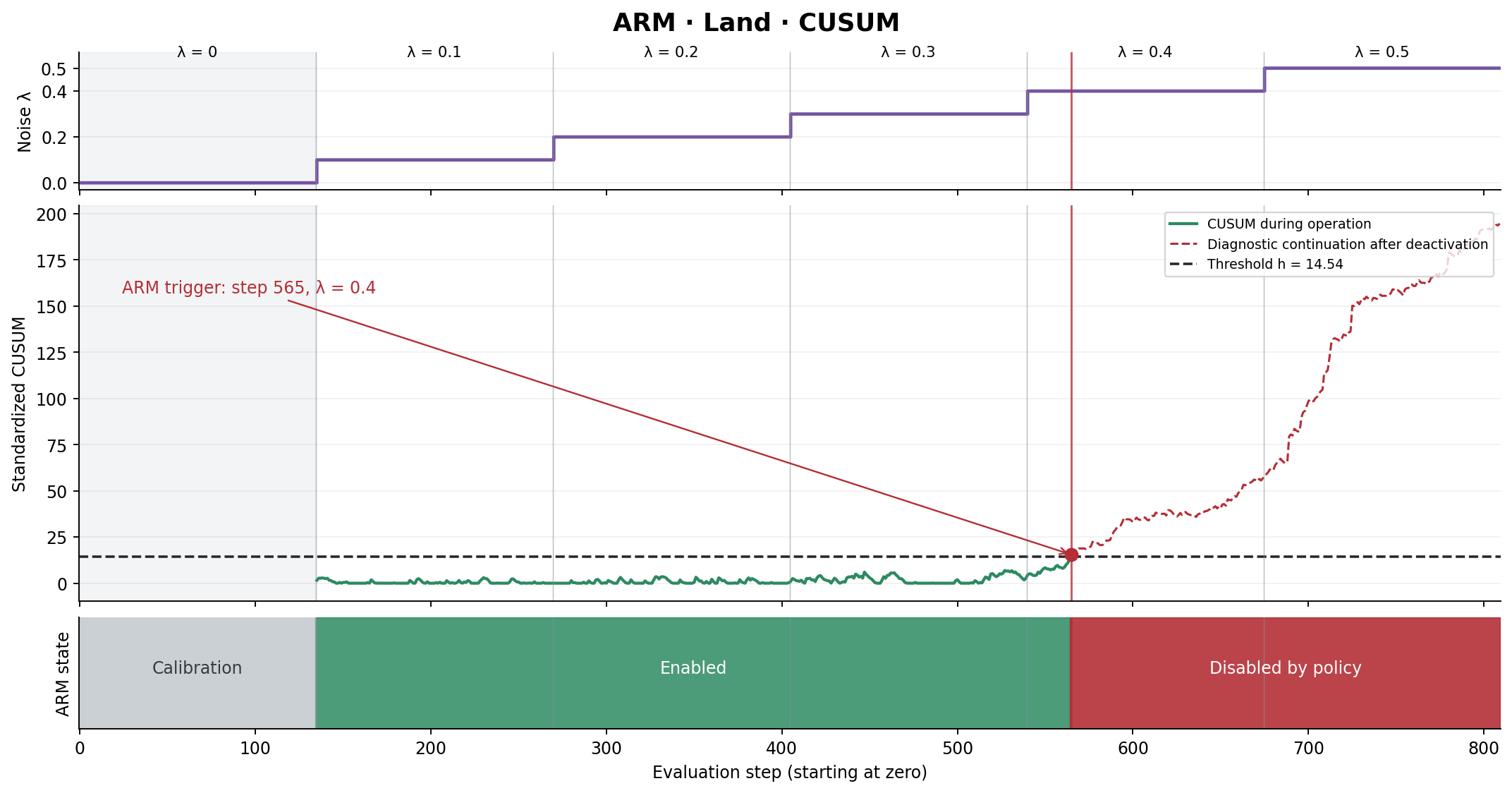}
  \caption{ARM monitoring for Land under increasing additive banding. The panels show banding strength, the CUSUM accumulator, and the operational state. The first crossing of $h\approx14.54$ at step 565 ($\lambda=0.4$) triggers deactivation. The attribute remains disabled thereafter; the red dashed curve shows diagnostic continuation of the accumulator.}
  \label{fig:arm_land}
\end{figure*}

Figure~\ref{fig:arm_land} illustrates the CUSUM evolution and the resulting operational state for \emph{Land}.

The three disabled attributes showed higher coverage MAE at the final noise level than under clean input. MAE increased from 2.68 to 20.31 percentage points for \emph{Ice/Snow}, from 18.34 to 25.06 for \emph{Land}, and from 4.82 to 6.53 for \emph{Cloud shadow}. However, deterioration did not always trigger an alarm: \emph{Clouds} remained enabled although its MAE increased from 12.82 to 14.99 percentage points. \emph{Water} and \emph{Cloud shadow over water} remained enabled and showed slightly lower final MAE.

\begin{table*}[ht]
\centering
\begin{tabular}{lcccc}
\hline
\textbf{Attribute} &
\textbf{Final state} &
\textbf{Initial MAE} &
\textbf{Final MAE} &
\textbf{$\lambda$ at alarm} \\
\hline
Water                   & Enabled & 6.03  & 5.64  & None \\
Clouds                  & Enabled & 12.82 & 14.99 & None \\
Ice/Snow                & Disabled & 2.68  & 20.31  & 0.2 \\
Land                    & Disabled & 18.34 & 25.06 & 0.4 \\
Cloud shadow            & Disabled & 4.82  & 6.53  & 0.5 \\
Cloud shadow over water & Enabled  & 0.69  & 0.66  & None \\
\hline
\end{tabular}
\caption{Final ARM state and coverage MAE in percentage points.
Initial and final MAE correspond to $\lambda=0$ and $\lambda=0.5$,
respectively. None indicates that no alarm was triggered.}
\label{tab:arm_results}
\end{table*}

\subsubsection{Gaussian Blur}

Under increasing Gaussian blur, ARM disabled \emph{Water} and \emph{Ice/Snow} at $\sigma=0.5$, \emph{Clouds} at $\sigma=1$, and \emph{Land} and \emph{Cloud shadow} at $\sigma=2$. Only \emph{Cloud shadow over water} remained enabled. As in the banding experiment, these values indicate when alarms occurred in the monitored sequence rather than fixed tolerance limits.

Four of the five disabled attributes showed higher final coverage MAE than under clean input. MAE increased from 6.03 to 9.74 percentage points for \emph{Water}, from 12.82 to 13.41 for \emph{Clouds}, from 18.34 to 19.85 for \emph{Land}, and from 4.82 to 5.32 for \emph{Cloud shadow}. In contrast, \emph{Ice/Snow} was disabled although its overall MAE decreased from 2.68 to 2.45 percentage points.

\begin{table*}[ht]
\centering
\begin{tabular}{lcccc}
\hline
\textbf{Attribute} &
\textbf{Final state} &
\textbf{Initial MAE} &
\textbf{Final MAE} &
\textbf{$\sigma$ at alarm (px)} \\
\hline
Water                   & Disabled & 6.03  & 9.74  & 0.5 \\
Clouds                  & Disabled & 12.82 & 13.41 & 1 \\
Ice/Snow                & Disabled & 2.68  & 2.45  & 0.5 \\
Land                    & Disabled & 18.34 & 19.85 & 2 \\
Cloud shadow            & Disabled & 4.82  & 5.32  & 2 \\
Cloud shadow over water & Enabled  & 0.69  & 0.70  & None \\
\hline
\end{tabular}
\caption{Final operational state and coverage MAE under Gaussian blur.
MAE is reported in percentage points. Initial and final values correspond
to $\sigma=0$ and $\sigma=4$ pixels, respectively.
None indicates that no alarm was triggered.}
\label{tab:arm_blur_results}
\end{table*}

\section{General Discussion}
\label{sec:discussion}

\subsection{Limitations}
The current implementation relies on several design choices that were tailored to the monitored VLM. Six behavioral probes were manually selected using prior knowledge of the model and its semantic attributes, with the aim of obtaining overlapping descriptions that could reveal disagreement. The idea is similar to cross-questioning: a detective or clinician may ask about the same event in different ways to check whether the answers remain consistent. However, these questions are not universal and would need to be adapted to other models or tasks. Future work should explore more systematic or model-agnostic ways of generating behavioral probes, or alternative black-box inconsistency detectors inspired by methods such as SelfCheckGPT.

As noted in \cite{bidirectional_detector}, a model may produce consistent but incorrect answers. The detector used in this study monitors increases in disagreement between responses, rather than
semantic correctness. It may therefore fail to raise an alarm when the responses share the same error. This limitation concerns the selected evidence source; other detectors could be incorporated
into ARM to address failures that do not produce disagreement.

Under the experimental policy, an attribute remains disabled after its first alarm; recovery and reactivation were not evaluated. An alternative implementation could allow reactivation when the inconsistency evidence returns to its reference level, for example after a temporary change in observation conditions. Reactivation would require a separate policy, since restored consistency alone would not establish recovered semantic accuracy.

Finally, ARM has so far been evaluated in a single use case using a relatively simple VLM implementation. The 135 validation patches originate from only 15 source scenes and overlap spatially, so they cannot be treated as independent scene observations. The results may differ across models, datasets, and application domains, as the type and frequency of inconsistencies depend on the model, its training data, the prompting strategy, and the monitored attributes. Further evaluation using independent scenes and different models and tasks is therefore needed to assess how well these findings generalize.

\subsection{Implications for Autonomous Agents}

\begin{figure*}[ht]
  \centering
  \includegraphics[width=\textwidth]{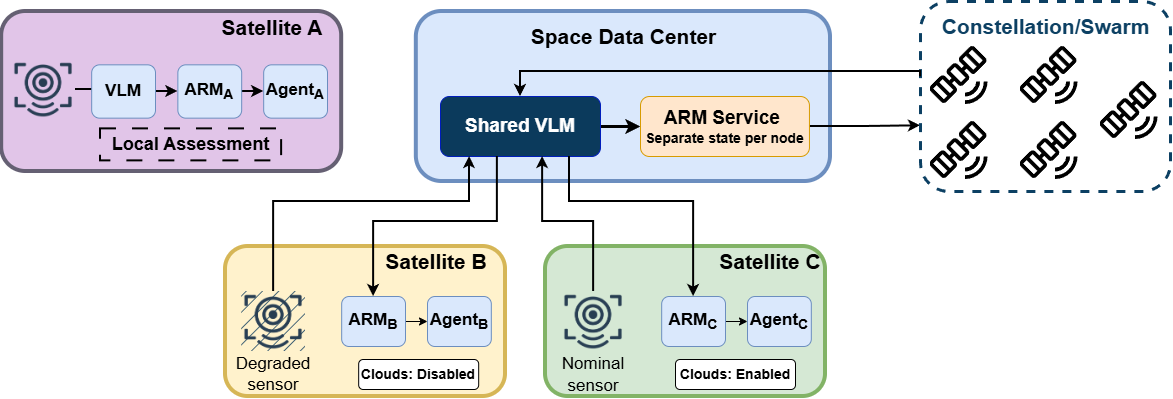}
  \caption{Conceptual deployment of ARM with local and shared models. Each agent maintains a separate attribute assessment associated with its operating conditions.}
  \label{fig:future_use}
\end{figure*}

Language and vision language models could become important components of future space missions with increasing autonomy. Their potential use extends beyond exploration missions to activities such as navigation support, instrument calibration, observation scheduling, and adaptation of processing tasks to mission needs or intermediate results. In constellations, these activities involve multiple platforms operating under different and changing conditions, making continuous human supervision increasingly difficult.

Such architectures would benefit from mechanisms that monitor the behavior of the models on which agents depend and help limit the consequences of unreliable outputs. ARM provides an initial framework for this purpose through external monitoring of individual semantic attributes. Its assessments could inform an agent's decision to use an alternative model or tool, request further verification, or restrict operations that depend on a suspended capability. These responses would be defined by the consuming system according to its operational requirements.

Conceptually, ARM could form a monitoring layer between the models and the agents that use their outputs, as illustrated in Fig.~\ref{fig:future_use}. This arrangement could apply both to models hosted onboard individual satellites and to shared services hosted in a space data center. Each agent could maintain an ARM assessment associated with its own instruments, tasks, and observation conditions. Two satellites accessing the same VLM could therefore maintain different operational states for the same semantic attribute if, for example, one instrument becomes degraded while the other remains nominal.

ARM itself could also be hosted as a shared service for multiple satellites. The service would maintain separate temporal memories, reference statistics, and attribute states for each satellite or monitoring context. This separation would preserve the history associated with each platform across successive requests, allowing a common service to provide individual assessments without pooling evidence from different operating conditions.

The present experiment demonstrates an initial implementation of this monitoring mechanism under controlled input degradation. Its integration into autonomous decision making and its evaluation across distributed platforms remain directions for future work.

\section{Conclusions and Future Work}
\label{sec:conclusions}

This work introduced the Attribute Reliability Monitor (ARM), a framework for accumulating evidence of suspicion over time and maintaining an operational assessment of individual semantic attributes produced by language and vision language models. ARM does not prescribe a particular detector, temporal monitoring method, or intervention policy, allowing its implementation to be adapted to the requirements of the consuming system.

To evaluate ARM, a specialized vision language model was developed for satellite thermal infrared imagery, using Landsat 8 Band 10 as its sole visual input, complemented by six auxiliary channels derived through physics inspired preprocessing. Monitoring followed a black box approach based only on observable responses, without access to internal model states, confidence scores, or external reference answers. Under controlled input degradation, the ARM implementation detected increases in disagreement between responses and supported selective suspension of individual attributes, while others remained enabled. These results illustrate how external behavioral evidence can inform decisions about which model outputs to continue using during operation. The assessment reflects accumulated suspicion and does not establish semantic correctness.

Future work should explore systematic generation of behavioral probes and complementary evidence sources for failures that do not produce disagreement. Evaluation across different models, tasks, and independent scenes would help assess the generality of the approach. Policies for reassessment and reactivation should also be investigated, alongside integration with autonomous agents that can select alternative models or tools when a monitored output becomes suspect.

For distributed space architectures, ARM could be deployed locally or provided as a shared service with separate temporal memories and reference conditions for each agent. This would allow agents accessing the same model to maintain assessments associated with their own instruments, tasks, and operating conditions. Evaluating these arrangements would help determine whether attribute monitoring can support continued operation under changing conditions.


\bibliographystyle{unsrt}
\bibliography{chapters/bibliography}

\small

\end{document}